\documentclass[review]{elsarticle}

\usepackage{lineno,hyperref}
\modulolinenumbers[5]

\journal{Expert Systems with Applications}

\biboptions{sort,authoryear}

\usepackage{booktabs,arydshln}

\usepackage{subcaption}
\usepackage{multirow}
\usepackage{ctable}
\usepackage{float}

\newcommand{\Fscore}{\text{F$_{1}$}}%
\newcommand{\Precision}{$\text{P}$}%
\newcommand{\Recall}{$\text{R}$}%
\newcommand{\MacroFscore}{\text{mF$_{1}$}}
\newcommand{\MacroPrecision}{\text{mP}}
\newcommand{\MacroRecall}{\text{mR}}

\newcommand{\Source}{$\mathcal{D}$}
\newcommand{\ImageI}{$\mathcal{I}$}

\newcommand{\Regions}{$\mathcal{R}$}
\newcommand{\DataAug}{$\mathcal{S}$}
\newcommand{\RegionsI}{$\mathcal{R}$}
\newcommand{\ImageSyn}{$\mathcal{I}_s$}
\newcommand{\RegionsSyn}{$\mathcal{R}_s$}
\newcommand{\RegionSyn}{$r_s$}
\newcommand{\region}{$r$}
\newcommand{\NumPages}{$n$}
\newcommand{\ImageRegion}{$i_s$}
\newcommand{\ImageSynInkCoords}{$\mathcal{C}_s$}
\newcommand{\ImageSynInkCoordX}{$x_s$}
\newcommand{\ImageSynInkCoordY}{$y_s$}
\newcommand{\RegionsCoordX}{$x_r$}
\newcommand{\RegionsCoordY}{$y_r$}
\newcommand{\MaxRotation}{$\Phi$}

\usepackage{bold-extra}
\newcommand{\Seils}{\textsc{Seils}}
\newcommand{\Capitan}{\textsc{Capitan}}

\newcommand{\FMTC}{\textsc{FMT-C}}
\newcommand{\FMTM}{\textsc{FMT-M}}
\newcommand{\FMT}{\textsc{FMT}}

\newcommand{\UniformRandom}{\emph{Our augmentation}}
\newcommand{\NoAugmented}{\emph{Non-augmented}}

\newcommand{\diffSER}{$\overline{\text{SER}}$}

\usepackage{acronym}
\acrodef{SAE}{Selectional Auto-Encoder}
\acrodef{CNN}{Convolutional Neural Network}
\acrodef{CRNN}{Convolutional Recurrent Neural Network}
\acrodef{TP}{True Positive}
\acrodef{FN}{False Negative}
\acrodef{FP}{False Positive}
\acrodef{OMR}{Optical Music Recognition}
\acrodef{OCR}{Optical Character Recognition}
\acrodef{FCN}{Fully-Convolutional Network}
\acrodef{GRL}{Gradient Reversal Layer}
\acrodef{GAN}{Generative Adversarial Network}
\acrodef{DA}{Domain adaptation}
\acrodef{DANN}{Domain-Adversarial Neural Network}
\acrodef{DNN}{Deep Neural Network}
\acrodef{R-CNN}{Region-based Convolutional Neural Network}
\acrodef{Fast R-CNN}{Fast Region-based Convolutional Neural Network}
\acrodef{Faster R-CNN}{Faster Region-based Convolutional Neural Network}
\acrodef{SSD}{Single Shot MultiBox Detection}
\acrodef{mAP}{mean Average Precision}
\acrodef{LA}{Layout Analysis}
\acrodef{ANN}{Artificial Neural Network}
\acrodef{SER}{Symbol Error Rate}
\acrodef{IoU}{Intersection over Union}

\newcommand{\RetinaNet}{\textsc{RetinaNet}}
\newcommand{\SSD}{\textsc{SSD}}
\newcommand{\FasterRCNN}{\textsc{Faster R-CNN}}
\newcommand{\SAE}{\textsc{SAE}}

\usepackage{amsfonts}
\usepackage{amsmath}
\usepackage{array}

\usepackage{algorithm}
\usepackage{algorithmicx}
\usepackage{algpseudocode}
\usepackage{setspace}

\algnewcommand{\LeftComment}[1]{\(\triangleright\) #1}

\usepackage{mathtools}% http://ctan.org/pkg/mathtools

\algnewcommand{\IfThenElse}[3]{% \IfThenElse{<if>}{<then>}{<else>}
  \State \algorithmicif\ #1\ \algorithmicthen\ #2\ \algorithmicelse\ #3}
\algnewcommand{\IfThen}[2]{% \IfThenElse{<if>}{<then>}
  \State \algorithmicif\ #1\ \algorithmicthen\ #2}
\algnewcommand{\VarIfThenElse}[4]{% \IfThenElse{<if>}{<then>}{<else>}
  \State #1\ $\gets$ \algorithmicif\ #2\ \algorithmicthen\ #3\ \algorithmicelse\ #4}
\algnewcommand{\VarIfThen}[3]{% \IfThenElse{<if>}{<then>}
  \State #1\ $\gets$ \algorithmicif\ #2\ \algorithmicthen\ #3}
\algnewcommand\algorithmicforeach{\textbf{for each}}
\algdef{S}[FOR]{ForEach}[1]{\algorithmicforeach\ #1\ \algorithmicdo}

\makeatletter
\def\adl@drawiv#1#2#3{%
        \hskip.5\tabcolsep
        \xleaders#3{#2\@tempdimb #1{1}#2\@tempdimb}%
                #2\z@ plus1fil minus1fil\relax
        \hskip.5\tabcolsep}
\newcommand{\cdashlinelr}[1]{%
  \noalign{\vskip\aboverulesep
           \global\let\@dashdrawstore\adl@draw
           \global\let\adl@draw\adl@drawiv}
  \cdashline{#1}
  \noalign{\global\let\adl@draw\@dashdrawstore
           \vskip\belowrulesep}}
\makeatother

\begin{document}

\begin{frontmatter}

%\title{Elsevier \LaTeX\ template\tnoteref{mytitlenote}}
\title{Region-based Layout Analysis of Music Score Images}



%\author[fcastellanos@dlsi.ua.es]{Francisco J. Castellanos}

\author[mymainaddress]{Francisco J. Castellanos\corref{mycorrespondingauthor}}
\cortext[mycorrespondingauthor]{Corresponding author}
\ead{fcastellanos@dlsi.ua.es}
\author[mymainaddress]{Carlos Garrido-Munoz}
\author[mymainaddress]{Antonio Ríos-Vila}
\author[mymainaddress]{Jorge Calvo-Zaragoza}

\address[mymainaddress]{Department of Software and Computing Systems, University of Alicante, Carretera San Vicente del Raspeig s/n, 03690, Alicante, Spain}

\begin{abstract}
The Layout Analysis (LA) stage is of vital importance to the correct performance of an Optical Music Recognition (OMR) system. It identifies the regions of interest, such as staves or lyrics, which must then be processed in order to transcribe their content. Despite the existence of modern approaches based on deep learning, an exhaustive study of LA in OMR has not yet been carried out with regard to the precision of different models, their generalization to different domains or, more importantly, their impact on subsequent stages of the pipeline. This work focuses on filling this gap in literature by means of an experimental study of different neural architectures, music document types and evaluation scenarios. The need for training data has also led to a proposal for a new semi-synthetic data generation technique that enables the efficient applicability of LA approaches in real scenarios. Our results show that: (i) the choice of the model and its performance are crucial for the entire transcription process; (ii) the metrics commonly used to evaluate the LA stage do not always correlate with the final performance of the OMR system, and (iii) the proposed data-generation technique enables state-of-the-art results to be achieved with a limited set of labeled data.
\end{abstract}


\end{frontmatter}

%\linenumbers

\section{Introduction}
\label{sec:introduction}
The digitization of music manuscripts helps preserve and disseminate this valuable heritage. However, simply obtaining a digital image from the original source is not sufficient to enable the computational use of this material, and it is, therefore, necessary to transcribe the content into a digital format.

The manual transcription of music sources is a time-consuming task. Given the countless number of music manuscripts scattered around the world, there are ongoing efforts to automatize this process by means of artificial intelligence. The research field that studies how to automatically transcribe music notation from scanned documents into a structured digital format is \ac{OMR}~\citep{Calvo-Zaragoza2019}. The structural complexity of music notation, along with the great variability as regards writing styles, engraving mechanisms or types of notation---such as neumatic, mensural or modern Western notation---makes the \ac{OMR} challenge far from straightforward.

Music manuscripts may contain not only music but also text, lyrics or document metadata, and it is for this reason that several tasks in the traditional \ac{OMR} workflow focus on the document itself with respect to the distribution of its different regions \citep{RebeloFPMGC12}. This is usually referred to as \ac{LA}, which is also common in other document contexts such as text recognition \citep{binmakhashen2019document}. The main purpose of \ac{LA} is to identify the relevant information from the whole image, thus facilitating the eventual objective of these systems: the transcription of their content. This can be done using two broad possible strategies \ac{LA}: pixel-wise or region-based analysis, the latter of which is the focus of this work.

Common formulations consider region-based \ac{LA} as an object detection task, in which each meaningful element or region of interest is located and classified into a set of predefined categories \citep{liu2020deep}. However, despite its importance in the \ac{OMR} context, and although literature contains a large number of general-purpose object-detection approaches, no comprehensive study has been performed in order to assess the behavior of different models for \ac{LA} when applied to music scores. In this work, we aim to fill this gap in literature by performing thorough experiments with several object-detection approaches that are evaluated in different scenarios.

Furthermore, the state of the art as regards \ac{LA} involves the use of machine learning, and particularly deep learning techniques~\citep{lecun2015}. The excellent performance demonstrated in several computer vision contexts makes this type of techniques appropriate for the task discussed herein.
However, the application of these techniques to \ac{LA} is not yet straightforward in real-case scenarios, since one of the major challenges is the need for sufficiently representative ground-truth data. This issue is particularly relevant in the context of music documents, given that manuscripts are highly heterogeneous and that reusing data from different collections is, therefore, ineffective. We propose to address this common issue through the use of an algorithm with which to generate semi-synthetic images in order to increment the set of available annotated data to be used as a reference for the deep learning approach.

Moreover, it should be noted that no specific metric is suitable for \ac{OMR} in order to evaluate the performance of \ac{LA}.One of the most common metrics used for assessment In the object-detection field is the \ac{mAP} metric, in which has also been employed the context of music~\citep{pacha2018baseline,jia2021printed}. However, no analysis has yet been carried out regarding whether it is indeed an appropriate metric that correlates with the quality of the bounding boxes extracted in order to eventually transcribe music. We, therefore, discuss the results obtained with different metrics in order to be able to state which is the most suitable for \ac{LA} when applied in \ac{OMR}.

Finally, since \ac{LA} is one of the earliest steps in the \ac{OMR} workflow, any inaccuracies might be critical as regards the eventual transcription, signifying that it is crucial to study its influence on the whole process. Analyzing the interaction between \ac{LA} and the further transcription is essential as regards stating the most appropriate way in which to address the \ac{OMR} process. However, \ac{LA} has usually been evaluated as an individual task without properly analyzing this question, and no existing study covers it. 

%For all the above reasons, in this work we: i) carry out exhaustive experiments and analyses of a series of object-detection models, selected from the state of the art as regards the general-purpose object retrieval field so as to cover multiple strategies; ii) analyze the behavior of the models in different scenarios according to the number of annotated pages available owing to the limited amount of annotated data that must be learned in typical OMR use-cases, and iii) propose a new data-generation mechanism for \ac{LA} with the aim of increasing the amount and variety of annotated images. %The idea behind this strategy is to train object-detection models that will be sufficiently robust to enable the extraction of regions of interest, despite the scarcity of reference data. 

To summarize, the contribution of this paper can be divided into the following points:
\begin{itemize}
    \item Carrying out the first comprehensive study of object-detection models for \ac{LA} in music score images.
    \item Analyzing and discussing the correlation between the common metrics used in object detection and the quality of the bounding boxes retrieved in the \ac{LA} process for \ac{OMR}.
    \item Proposing a new semi-synthetic data generation method for \ac{LA}, in addition to carrying out a thorough study of its usefulness.
    \item A goal-directed analysis of the influence of \ac{LA} on the eventual transcription.
\end{itemize}

The remainder of this paper is organized as follows: the state of the art of \ac{LA} is detailed in Section~\ref{sec:bg}, while the different architectures considered and our data-augmentation mechanism for \ac{LA} are described in Section~\ref{sec:method}. The experimental setup, along with the corpora and metrics considered, are explained in Section~\ref{sec:setup}, and Section~\ref{sec:results} shows the results obtained after carrying out staff-retrieval and end-to-end recognition experiments, in addition to the corresponding analysis and discussion of them. Finally, the main conclusions of the work are summarized in Section~\ref{sec:conclusions}.

\section{Background}
\label{sec:bg}

With regard to \ac{LA} in the context of \ac{OMR}, there are, in broad terms, two main perspectives that can be considered: processing the document at pixel level or at region level. 

The former perspective was traditionally addressed through the use of different strategies that can be found in literature. Before deep learning techniques were applied, other conventional systems were employed by means of heuristic techniques. For example, with regard to separately extracting the staff lines and lyrics from sheet music, \citet{burgoyne2009lyric} proposed a heuristic method with which to detect the waved text and staff lines that was based on the Hough transform. Although the staff lines are highly necessary as regards recognizing the pitch of the symbols, many \ac{OMR} workflows are based on this process, which is employed to perform a connected component analysis of the remaining music notation. There is, in this respect, a review that shows the earlier methods used for staff removal~\citep{dalitz2008comparative}, but new techniques have also been developed in order to address this question through the use of heuristic methods~\citep{dos2009staff,geraud2014morphological}. Of the topics related to \ac{OMR}, there is, among others, a review~\citep{RebeloFPMGC12} that gathers this type of solutions together in order to perform \ac{LA}. %\cite{gallego2017staff}

Despite the fact that these heuristic strategies may obtain good results in controlled scenarios, they are poorly generalizable, signifying that these methods are not, in practice, suitable for the processing of scanned documents. The major challenge in this respect is the great variability of this type of images owing to multiple factors, e.g. the degree of degradation, contrast, the color of the ink employed or skew variations, thus making it a difficult task to perform. The main focus as regards obtaining more generalizable models has been machine learning techniques, and particularly deep learning techniques. For example, \citet{calvo2018deep} presented a \ac{CNN}-based architecture with which to perform \ac{LA} by classifying each pixel of the image according to a set of categories. However, the method takes a long time because it processes each pixel in the image. In order to address this time-consuming issue, image-to-image strategy was also proposed~\citep{Castellanos2018ISMIR}, which is based on a series of encoder-decoder architectures (the so-called \ac{SAE}) and trains each one so as to extract a particular information layer. 

The region-wise perspective can, meanwhile, be considered as an object-detection process in which the objects are different parts of the document, such as staff regions or lyrics. Several previous works have followed this approach. One of the first was that of \citet{CamposCTV16}, which used Hidden Markov Models to carry out \ac{LA} in order to extract text and staff regions from music score images, while \citet{quiros2019multi} proposed the use of an \acl{ANN} architecture to extract the different regions of interest from a music document. In their work, \citet{pacha2019incremental} developed an incremental method for the training of supervised models with a combination of annotated and predicted images so as to extract the bounding boxes of staves. Moreover, \citet{Waloschek2019} proposed a neural network approach that could be used to extract the bounding boxes of the system measures from a music score image, focusing on the alignment between them. A full-page framework based on two steps---staff recognition and end-to-end transcription---was recently proposed~\citep{CastellanosIsmir2020}. With respect to the first step, it performed \ac{LA} in order to extract the staff regions by means of an \ac{SAE} and connected-component analysis. In their work, meanwhile, \citet{marc2021} proposed the use of \ac{Faster R-CNN} to detect the bounding boxes of staves and measures. 

In addition, note that a large number of region-based \ac{LA} methods use \ac{mAP} as a metric to evaluate the quality of the bounding boxes~\citep{pacha2018baseline,Waloschek2019,marc2021,Huang2019}, but no study of the suitability of this metric for \ac{OMR} can be found in literature. 

%However, despite the close relationship between region-based \ac{LA} and object recognition, to the best of our knowledge no fitting study of the behavior of different object-detection models in the context of applying \ac{LA} in the \ac{OMR} context has as yet been carried out. There are, however, a large number of well-known approaches that perform object detection in \ac{OCR} well, and which could be potential solutions as regards processing music documents. For example, with regard to textual documents, \citet{soto2019visual} used \ac{Faster R-CNN} and ResNet to detect different areas by searching for titles, captions, paragraphs or figures, among others. The approach finds a list of regions of interest and a classifier determines the type of region to which they belong. Another example is that developed by \citet{ryu2019chinese}, in which Chinese characters are detected by means of \ac{SSD}, or that of \citet{sokerin2021object}, in which the use of RetinaNet is proposed in order to identify elements in financial reports. The relevance of \ac{LA} in textual documents signifies that literature even contains the description of a contest~\citep{icdar_layout}. 

\section{Methodology}
\label{sec:method}
This section provides a description of the methodology considered for \ac{LA}. It is divided into two parts: the description of the different object-detection models deemed appropriate for this task, and the definition of our data augmentation proposal with which to generate new semi-synthetic images, which is particularly useful when there is insufficient annotated data.

\subsection{Object-detection architectures for \acl{LA}}
\label{sec:method:architectures}

We considered several well-known general-purpose models for the task of applying object detection in \ac{LA}. These were selected owing to their popularity and considerable capabilities in multiple areas, in addition to the fact that they cover various neural strategies, such as one-stage or two-stage models or even pixel-wise segmentation. We particularly used those shown in the following list:

\begin{itemize}
    \item \FasterRCNN{}~\citep{ren2015faster} is a two-stage detection model that includes a Region Proposal Network (RPN) to Fast R-CNN~\citep{girshick2015fast}. This model uses the last convolutional layer of the backbone as a feature map and attempts to extract proposals for classification and localization directly through the RPN. These proposals are also used to train the classifier, enabling it to create a unified network. Since convolutional features are shared, the efficiency of training increases with regard to other previous architectures such as Fast R-CNN and R-CNN. This model has proven to be highly efficient and to perform well in several scenarios, thus making it an ideal candidate as regards carrying out \ac{LA}. We consider this detector alongside the ResNet50 backbone~\citep{ResNet}.

    \item \RetinaNet{}~\citep{RetinaNet}
    is a one-stage object-detection model composed of a backbone and two sub-networks. The backbone part computes the convolutional feature map over the input image, typically relying on ResNet and adopting the Feature Pyramid Network (FPN) in order to extract proposals. The sub-network part is composed of classification and box regression networks. RetinaNet attempts to solve the common problem of imbalanced data, in which there are different numbers of samples for each class. This issue usually causes a bias in the training by tending toward predicting the majority class (usually, ``background''). RetinaNet addresses this by means of a focal loss function, which dynamically shifts weights in order to decrease the contribution of well-classified samples and focuses on misclassifications by means of the modulating factor of the focal loss. This model is especially interesting for \ac{LA} since the content of a music document is very varied and may contain disparate elements, such as a different number of staff and text regions. We use this detector in combination with ResNet50 and FPN.  

    \item \SSD{}~\citep{SSD} 
    is a one-stage model that takes feature maps in order to generate multi-scale proposal predictions. It retrieves objects in one step and explicitly divides the predictions by employing an aspect ratio. We use this detector with VGG16~\citep{SimonyanZ14a} as a backbone, since it is the basis of the original work.

    \item \SAE{} is a \ac{FCN}, and specifically a U-net architecture~\citep{unet}, which is able to classify each pixel of an image according to a set of categories. This type of architecture is composed of two parts: an encoder that extracts the relevant features with combinations of convolutional and pooling layers, and a decoder that inverts the encoder operation with convolutional and up-sampling layers until the size of the input image is retrieved. The \ac{SAE} model provides a probabilistic map whose elements contain the probability of each pixel belonging to a specific class. This model has been successfully used for the staff-retrieval task with music score images~\citep{CastellanosIsmir2020}. It is important to highlight that, to be able to apply this method in \ac{LA}, a post-process is required in order to convert the probabilistic map obtained by the neural network into a set of bounding boxes by performing a connected-component analysis.

\end{itemize}

It should be noted that the three first architectures presented---\FasterRCNN{}, \RetinaNet{} and \SSD{}---rely on the use of a handcrafted technique called Non-Maximum-Supression~\citep{neubeck2006efficient} to deal with multiple detections of the same object. 

Moreover, the \SAE{} method has the restriction of being unable to detect overlapped bounding boxes. We have, therefore, followed the strategy of~\citep{CastellanosIsmir2020} so as to vertically reduce the bounding boxes by $20\%$ of the ground truth in order to mitigate this restriction. With regard to the predictions, after extracting the coordinates of the bounding boxes, the method vertically increases the predicted bounding boxes by the same ratio. Note that this alteration is not necessary for the other methods. In addition, \ac{SAE} does not provide a confidence value for each precision, which is the degree of certainty that the model has for its estimation, although the other models do provide it.

\subsection{Semi-synthetic data generation for \acl{LA}}
\label{sec:method:algorithm}

Let us consider \Source{}, a collection of labeled images consisting of pairs (\ImageI{}, \RegionsI{}), in which \ImageI{} is an image and \RegionsI{} is its respective ground-truth bounding boxes or regions. In this work, global coordinates are used to define each of these regions with its location within \ImageI{} and the class to which it belongs---\texttt{staff} or \texttt{text}, but more classes could be applicable. The idea behind our data-augmentation algorithm is to take advantage of the often scarce ground-truth data available in the \ac{OMR} context in order to build new semi-synthetic images composed of a combination of individual elements extracted from the original images.

\begin{algorithm*}[!ht]
% \setstretch{1.5}
    \caption{Image-generation algorithm proposed.}
    \label{alg:data_aug}
    \begin{algorithmic}[1] 
        \Function{Image-Generation}{\Source{}, \NumPages{}, \MaxRotation{}} 
        \State \DataAug{} $\gets \emptyset$
        \For{$i \gets 1$ to \NumPages{}} 
            \State \ImageI{}$, \, $\RegionsI{} $\gets$ {\it image-selection-policy}(\Source{})
            \State \ImageSyn{} $\gets$ {\it background-estimation}(\ImageI{})
            \State \RegionsSyn{} $\gets \emptyset$
            \ForEach {\region{} $\in$ \RegionsI{}} 
                \State \RegionSyn{}, \ImageRegion{} $\gets$ {\it region-selection-policy}(\region{}, \Source{})
                \State \RegionsCoordX{}, \RegionsCoordY{} $\gets$ {\it reference-global-coordinates}(\region{})
                \State \RegionSyn{} $\gets$ {\it update-global-coordinates}(\RegionSyn{}, \RegionsCoordX{}, \RegionsCoordY{})
                \State \RegionSyn{}, \ImageRegion{} $\gets$ {\it distortion-policy}(\RegionSyn{}, \ImageRegion{}, \MaxRotation{})
                \State \RegionsSyn{} $\gets$ \RegionsSyn{}\, $\cup\,$ \RegionSyn{}
                \State \ImageSynInkCoords{} $\gets$ {\it ink-detection}(\ImageRegion{})
                \ForEach {(\ImageSynInkCoordX{}, \ImageSynInkCoordY{}) $\in$ \ImageSynInkCoords{}} 
                    \State \ImageSyn{}[\RegionsCoordX{} + \ImageSynInkCoordX{}][\RegionsCoordY{} + \ImageSynInkCoordY{}] $\gets$ \ImageRegion{}[\ImageSynInkCoordX{}][\ImageSynInkCoordY{}]
                \EndFor
            \EndFor
            \State \DataAug{} $\gets$ \DataAug{}\, $\cup\,$ (\ImageSyn{}, \RegionsSyn{})
        \EndFor
		\State \Return \DataAug{} 
        \EndFunction
    \end{algorithmic}
\end{algorithm*}

The proposed data-augmentation mechanism is described in Algorithm~\ref{alg:data_aug}. The principal idea is to build a new dataset \DataAug{} from \Source{} with $n$ generated images. \DataAug{} consists of a set of pairs in the form of (\ImageSyn{}, \RegionsSyn{}), in which \ImageSyn{} is a semi-synthetic image and \RegionsSyn{} is the respective ground-truth data for the bounding boxes of \ImageSyn{}. 

In order to obtain a realistic image \ImageSyn{}, the algorithm first selects an existing image \ImageI{} and its respective ground-truth data \RegionsI{} from \Source{} by means of the function {\it image-selection-policy}$(\cdot, \cdot)$, as shown in line 4. In this work, this selection was made randomly, but other policies could also be applied. 

The next step consists of generating a background for the new image. This is performed in line 5 by the function {\it background-estimation}$(\cdot)$, which applies a process to the selected image \ImageI{} in order to build the basis of the new image. We propose using a blurring operation, with the objective of fading the content of the image and obtaining an empty image with a similar background to that of the original one.

Once the background has been built, and in order to keep the structure of the original image \ImageI{}, we propose replacing each region \region{} $\in$ \RegionsI{} with another same-class region \RegionSyn{} that is available in \Source{}. For example, if \region{} represents a bounding box of a music staff, the new region \RegionSyn{} will also be a staff region selected from all those available in \Source{} in order to then locate them in the same position as \region{}. This selection is carried out in line 8 by the function {\it region-selection-policy}$(\cdot, \cdot)$, which applies a selection policy in order to search for a replacement for each \region{} $\in$ \RegionsI{}. In our case, we consider a random selection of same-class regions. Note that in addition to extracting the new region \RegionSyn{}, the portion of image~\ImageRegion{} represented by the bounding box is also extracted. 

Since \RegionSyn{} contains coordinates relative to the original image from which \RegionSyn{} was extracted, it is necessary to adjust the coordinates to those relative to the new image. The method, therefore, extracts the reference coordinates \RegionsCoordX{}, \RegionsCoordY{} from \region{} by means of the function {\it reference-global-coordinates}$(\cdot)$ shown in line 9. In our case, we use the upper-left corner of the bounding boxes as reference coordinates. These coordinates are then used to update the coordinates of the new bounding box \RegionSyn{} by means of the function {\it update-global-coordinates}$(\cdot, \cdot, \cdot)$, which is in line 10. 

For the sake of more variability, in line 11, the function {\it distortion-policy}$(\cdot, \cdot, \cdot)$ applies a distortion policy \MaxRotation{} to \ImageRegion{} as an image-augmentation process. We apply a slight random rotation with respect to the center of each region, of between -3º and 3º--- a range used in previous work~\citep{juanki2021} for data augmentation in \ac{OMR}---with respect to the original skew, and this is the same value for all the regions on a page, but is different for other pages. It should be noted that excessive rotation could lead to the overlapping of multiple bounding boxes, which would lead to the attainment of unrealistic images. This function also updates the coordinates of \RegionSyn{} according to the distortion applied. \RegionSyn{} must subsequently be included in \RegionsSyn{}, as stated in line 12.

\begin{figure}[!ht]
    \centering
    \includegraphics[width=0.8\textwidth]{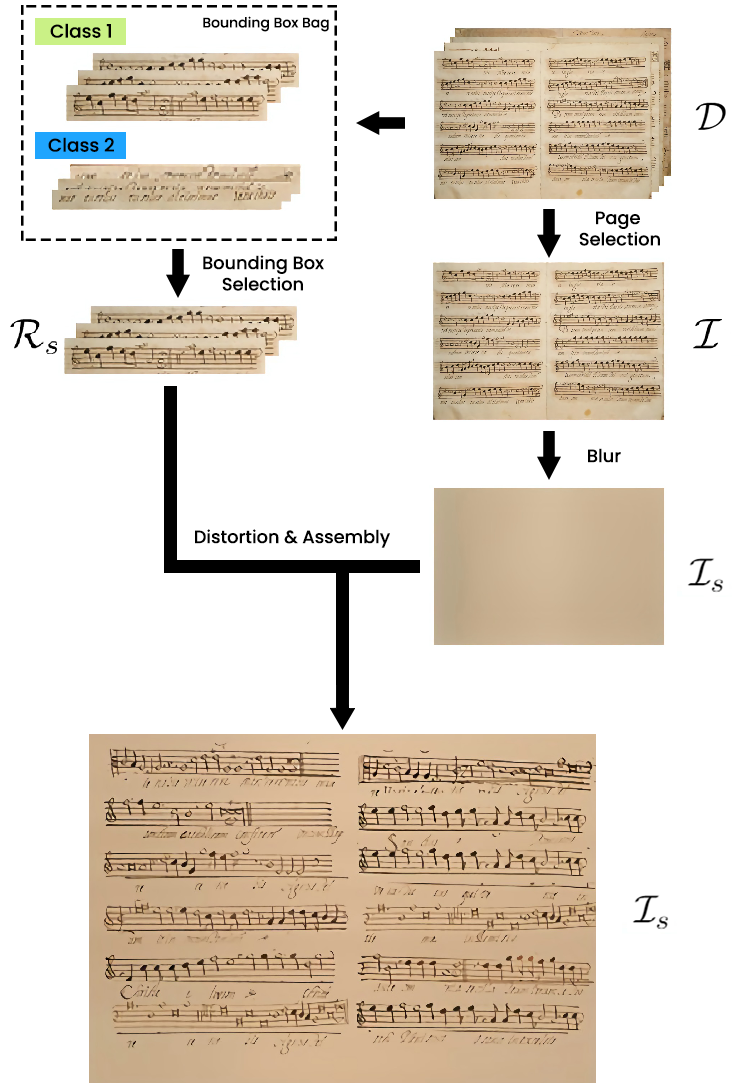}
    \caption{Overview of the data-generation algorithm proposed.}
    \label{fig:pipeline_generator}
\end{figure}

At this point, the algorithm must dump the content of the new bounding box \ImageRegion{} onto the image generated \ImageSyn{}. However, it is important to emphasize that, although the background is obtained by processing a real image, it is not exactly the same as that in the original images. We, therefore, propose using only the relevant information from \ImageRegion{}---the pixels with ink---and avoiding the background pixels. In order to perform this, in line 13, the function represented as {\it ink-detection}$(\cdot)$ returns the set of coordinates~\ImageSynInkCoords{} in the form (\ImageSynInkCoordX{}, \ImageSynInkCoordY{}), which represent the relative positions of the ink within \ImageRegion{}. Literature contains countless binarization approaches that can be used for this purpose~\citep{pastor-pellicer15insights,he2019deepotsu}. In this work, we applied the well-known local-thresholding algorithm for binarization developed by \citet{sauvola2000adaptive}, but any other could be used. The ink pixels of \ImageRegion{}, which are indicated by the relative coordinates \ImageSynInkCoords{}, are then dumped onto \ImageSyn{} by using \RegionsCoordX{}, \RegionsCoordY{} to properly locate the ink information. In the algorithm, this is performed in lines 14-16.

Finally, once the above process has been completed for all the regions in \Regions{}, the new semi-synthetic image \ImageSyn{} and its respective ground-truth data \RegionSyn{} are included in \DataAug{}, as stated in line 18, which contains the augmented dataset that will be returned at the end of the algorithm. The entire process is then repeated until \NumPages{} new images have been generated. The algorithm described is also shown schematically in Figure~\ref{fig:pipeline_generator}.

It should also be noted that, although the documents belongs to the same manuscript, the same-class regions of crossing pages may be of different sizes. We, therefore, considered skipping those replacements in which the inclusion of \RegionSyn{} within \RegionsSyn{} causes overlapping between different bounding boxes.

\section{Experimental Setup}
\label{sec:setup}

\subsection{Corpora}
\label{sec:setup:corpora}
\begin{figure*}[!ht]
     \centering
     \begin{subfigure}[b]{0.45\textwidth}
         \centering
         \includegraphics[width=\textwidth]{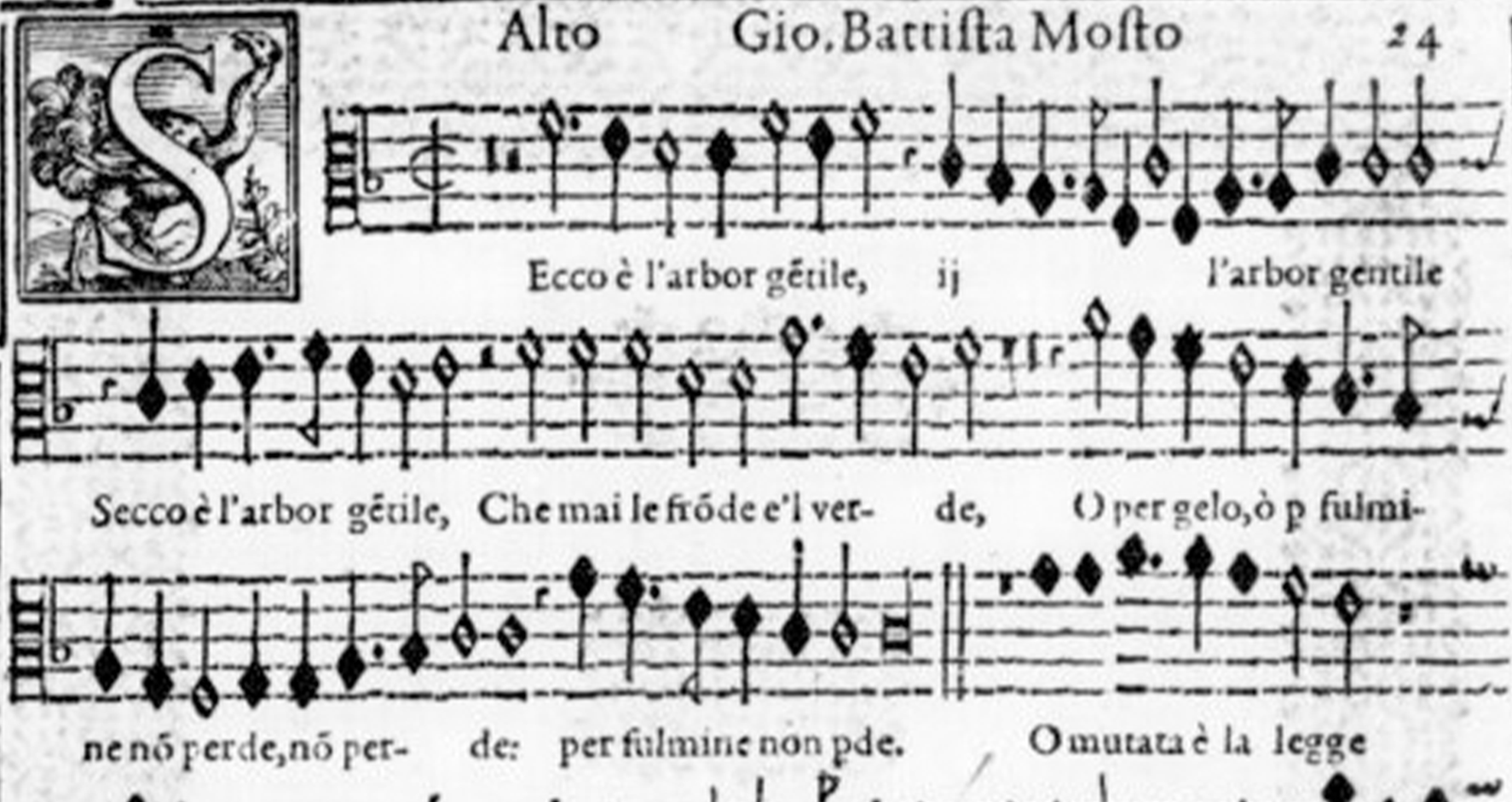}
         \caption{\Seils{}}
         \label{fig:seils}
     \end{subfigure}
     \begin{subfigure}[b]{0.45\textwidth}
         \centering
         \includegraphics[width=\textwidth]{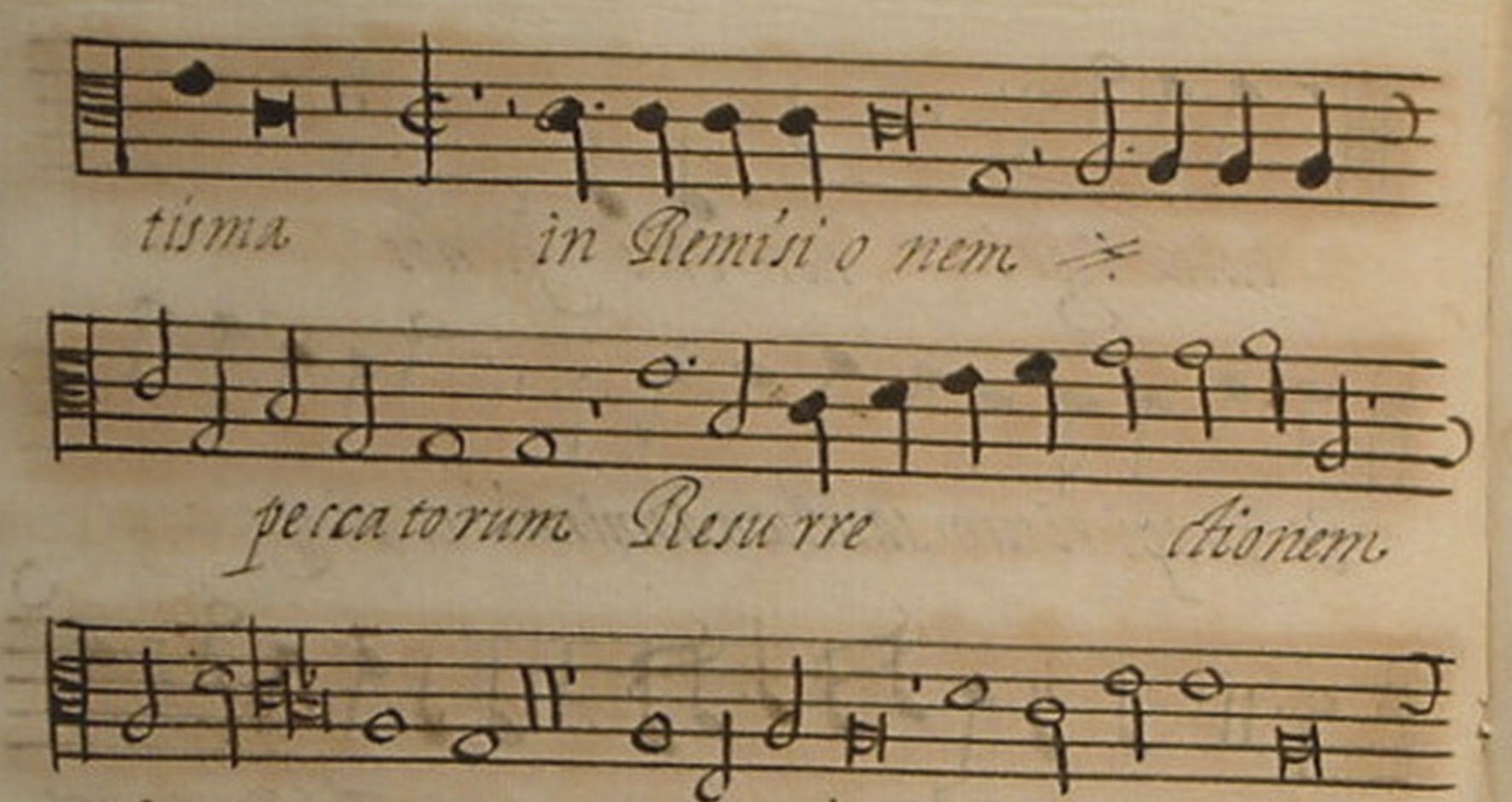}
         \caption{\Capitan{}}
         \label{fig:capitan}
     \end{subfigure}
     \begin{subfigure}[b]{0.45\textwidth}
         \centering
         \includegraphics[width=\textwidth]{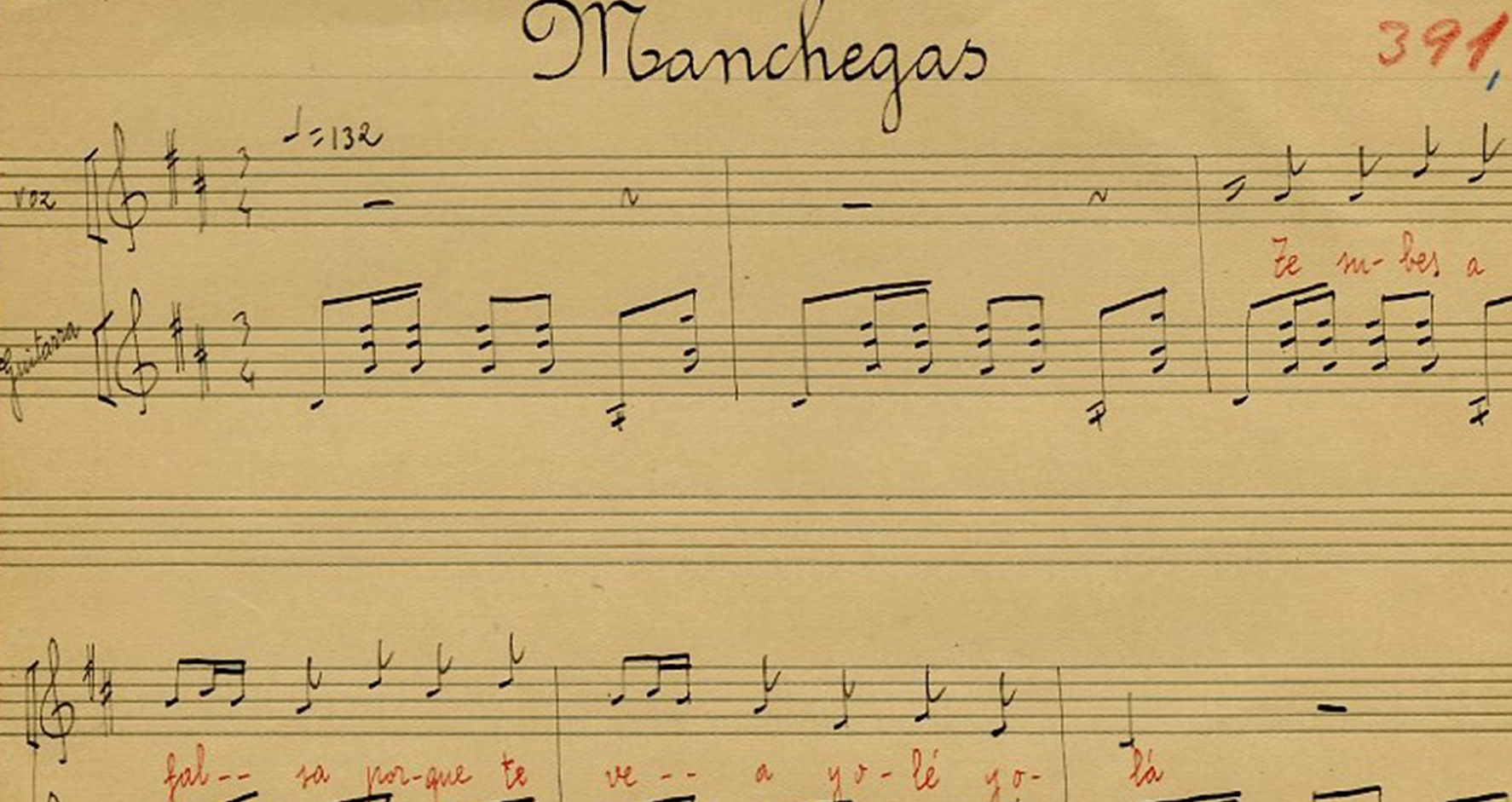}
         \caption{\FMTM{}}
         \label{fig:fmtm}
     \end{subfigure}
     \begin{subfigure}[b]{0.45\textwidth}
         \centering
         \includegraphics[width=\textwidth]{img/fmt-m.png}
         \caption{\FMTC{}}
         \label{fig:fmtc}
     \end{subfigure}
     \caption{Samples of the corpora considered for the experimentation.}
     \label{fig:corpora}
\end{figure*}

With regard to the experimentation, we considered several music corpora. These were selected because of their dissimilar nature, as depicted in Figure~\ref{fig:corpora}, in order to attain a better understanding of the behavior of the proposed methodology depending on the challenge. We specifically considered the following corpora, whose details are shown in Table~\ref{tab:corpora_characterization}:
\begin{itemize}
    \item \Seils{}: This dataset contains 150 typeset pages of the Il Lauro Secco manuscript~\citep{parada2019diplomatic} corresponding to an anthology of 16th-century Italian madrigals in mensural notation.
    \item \Capitan{}: This corpus is a compilation of 17th and 18th century manuscripts from the `Cathedral of Our Lady of the Pillar' in Zaragoza (Spain)\footnote{RISM Code `E-Zac` at https://rism.info/}. This dataset is an evolution of the `Zaragoza` corpus, which was created manually and introduced by~\citet{CalvoZaragoza:2016vc}.
    \item \FMT{}: The `Fondo de Música Tradicional IMF-CSIC` corpus~\citep{FabregasFMT:2021} consists of a collection of four groups of handwritten score sheets for popular Spanish songs transcribed by musicologists between 1944 and 1960. As it contains various manuscripts with dissimilar features, such as page color, image resolution or staff-region size, among others, these manuscripts have been clustered by similarity into two datasets: \FMTM{} and \FMTC{}, whose graphic differences are depicted in Figures~\ref{fig:fmtm} and~\ref{fig:fmtc}, respectively. 
\end{itemize}

\begin{table}[!h]
\centering
\caption{Description of the corpora. The ``Descr.'' column represents the description. In addition, in the ``Engraving'' row, ``Hw.'' signifies handwritten pages, whereas ``Pr.'' represents printed ones.}
\renewcommand{\arraystretch}{0.75}
\centering
\resizebox{0.75\textwidth}{!}{
\begin{tabular}{@{}rlrrrrr@{}}
\toprule
& \textbf{Descr.} & \multicolumn{1}{c}{\textbf{\Seils{}}} & \multicolumn{1}{c}{\textbf{\Capitan{}}} & \multicolumn{1}{c}{\textbf{\FMTM{}}} & \multicolumn{1}{c}{\textbf{\FMTC{}}} & \\ 
\cmidrule(rl){1-7}
& Engraving & Pr. &Hw. &Hw. &Hw. & \\
&Pages  & $150$ & $96$ & $703$ & $140$ & \\
& Lyrics & $2\,237$ & $695$ & $1\,241$ & $452$ & \\
& Staves & $1\,430$ & $775$ & $1\,508$ & $1\,435$ & \\
& Symbols & $31\,589$ & $17\,115$ & $11\,327$ & $5\,766$ & \\
\bottomrule
\end{tabular}}
\label{tab:corpora_characterization}
\end{table}

It should be noted that, when evaluating \ac{LA}, we study the behavior of the object-detection models in situations in which a different number of annotated images is used for the training process. In order to perform the same experiments for all corpora, we, therefore, fixed a maximum of 64 pages for the training of the models, increasing from 1 to 64 in powers of two. The other pages were equally divided into validation and testing partitions.

\subsection{Metrics}
\label{sec:setup:metrics}
The proposed methodology was evaluated by considering different metrics, according to the specific experiment being carried out. 

With regard to the \ac{LA} experiments, we considered the COCO \ac{mAP} metric \citep{lin2014microsoft}, which is widely used to evaluate object-detection models. This metric computes the area over the precision-recall curve, considering a range of values of \ac{IoU}, ranging from 0.5 to 0.95 in intervals of 0.05. 

However, the importance of \ac{LA} in \ac{OMR} lies mainly in the region retrieval and not so much in how well the predictions fit the ground-truth bounding boxes. As will be discussed at greater length in Section~\ref{sec:results:second_experiment}, this signifies that \ac{mAP} is not an appropriate metric for \ac{LA}. We shall, therefore, also evaluate \ac{LA} in terms of precision \Precision{}, recall \Recall{} and the harmonic mean F-score (\Fscore{}), which are computed as follows:

\begin{equation}
\mbox{P}= \frac{\mbox{TP}}{\mbox{TP} + \mbox{FP}},~~~\mbox{R}= \frac{\mbox{TP}}{\mbox{TP} + \mbox{FN}},
\label{eq:precision}
\end{equation}

\begin{equation}
\mbox{F}_{1} = 2\cdot \frac{\mbox{P} \cdot \mbox{R} }{\mbox{P} + \mbox{R}},
\label{eq:f1}
\end{equation}

\noindent where TP, FP, and FN, in our context, represent {\it True Positives} or correctly classified regions, {\it False Positives} or type I errors refer to those predictions that do not match a real bounding box, and {\it False Negatives} or type II errors refer to the real regions that have not been detected, respectively. Note that these metrics are evaluated and computed with respect to one class. For the evaluation of multiple classes through the use of a single value, these metrics can instead be reformulated as the macro average, in which  macro-precision, macro-recall and macro-\Fscore{}---henceforth \MacroPrecision{}, \MacroRecall{} and \MacroFscore{} are, respectively---the average of~\Precision{}, \Recall{} and \Fscore{} for all the classes involved.

We shall also evaluate the quality of the regions detected in terms of the ability of a state-of-the-art \ac{OMR} model to retrieve the musical symbols from them. In this case, the effectiveness of the transcription system is typically measured using the \ac{SER} metric \citep{Calvo-ZaragozaT19}. Let $H$ be the hypothesized sequence of music symbols and $R$ be the ground-truth sequence, and let \ac{SER} be computed by dividing the Levenshtein distance between $H$ and $R$ by the length of $R$.

\section{Results}
\label{sec:results}
In this section, we analyze the results obtained after carrying out three case studies for which the means employed to analyze the performance or the goal of the experiment was varied: (i) a standard evaluation, in which the typical metric in object detection (\ac{mAP}) was used to assess the predicted bounding boxes; (ii) an evaluation in terms of retrieved regions, in which the estimations were assessed by employing \Precision{}, \Recall{} and \Fscore{}, thus emphasizing the retrieval of bounding boxes rather than \ac{IoU}; and finally, (iii) a goal-directed evaluation in which a study of the influence of the \ac{IoU} and the confidence provided by the \ac{LA} model in the final transcription---scored by means of \ac{SER}---is discussed.

\subsection{Case Study I: Standard evaluation}
\label{sec:results:first_experiment}

In this section, we present the results obtained after experimenting with a series of object-detection models whose purpose is to perform the \ac{LA} of music score images at the region level, as described in Section~\ref{sec:method:architectures}. 

\begin{figure*}[!ht]
     \centering
     \begin{subfigure}[b]{0.99\textwidth}
         \centering
         \includegraphics[width=\textwidth]{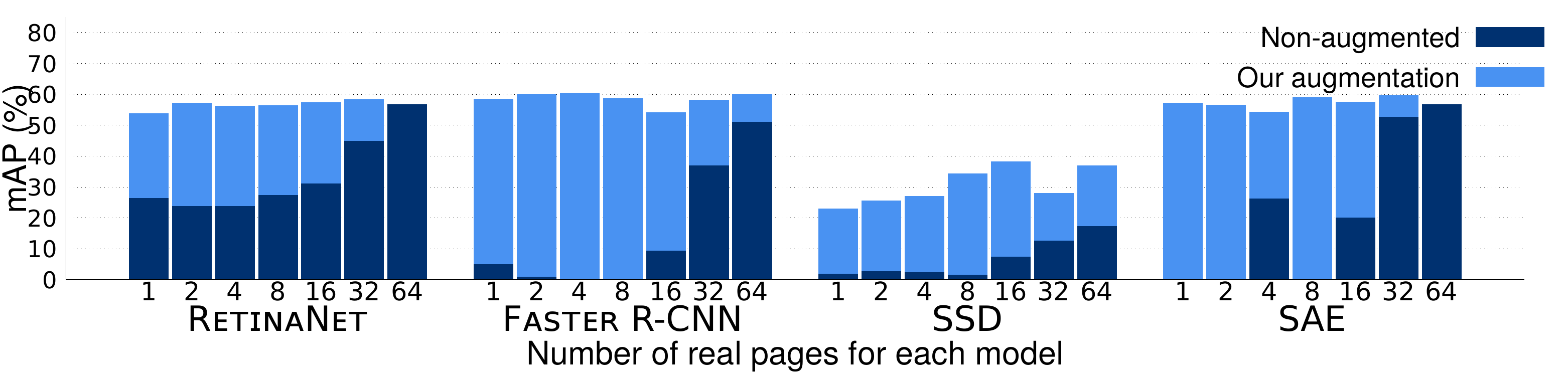}
         \caption{\Capitan{}}
         \label{fig:map_capitan}
     \end{subfigure}
     \begin{subfigure}[b]{0.99\textwidth}
         \centering
         \includegraphics[width=\textwidth]{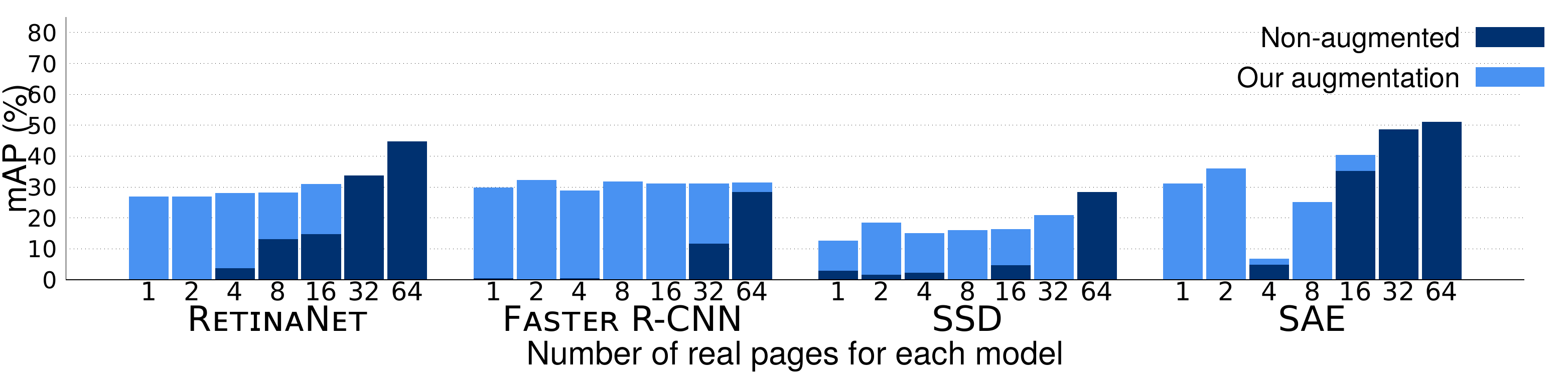}
         \caption{\Seils{}}
         \label{fig:fig:map_seils}
     \end{subfigure}
     \begin{subfigure}[b]{0.99\textwidth}
         \centering
         \includegraphics[width=\textwidth]{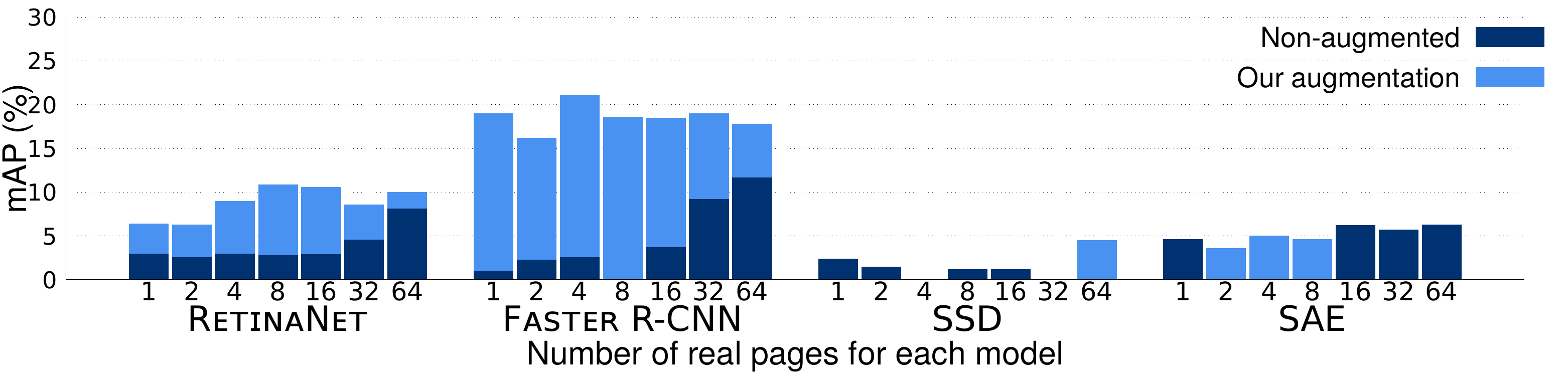}
         \caption{\FMTC{}}
         \label{fig:map_fmtc}
     \end{subfigure}
     \begin{subfigure}[b]{0.99\textwidth}
         \centering
         \includegraphics[width=\textwidth]{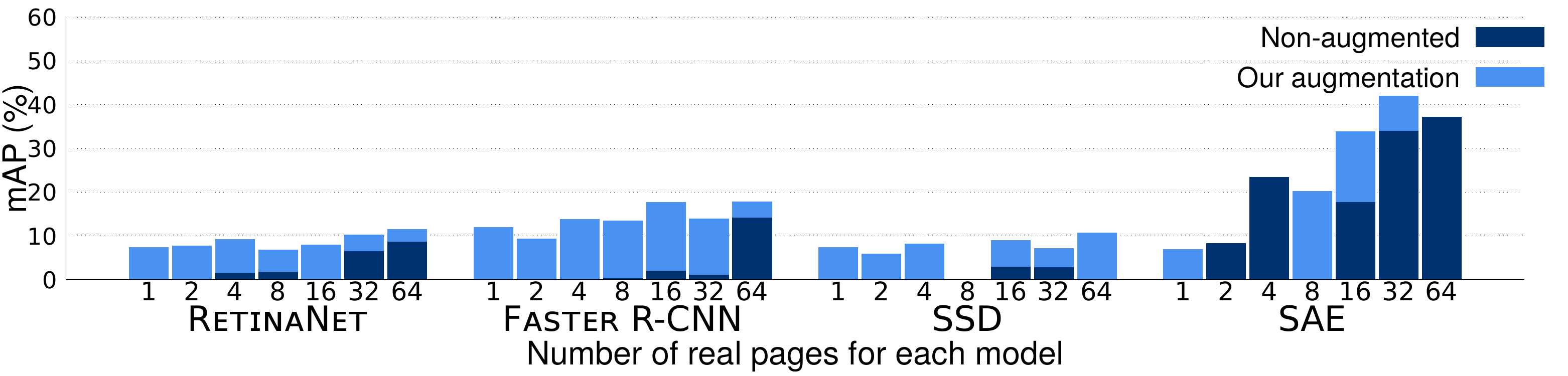}
         \caption{\FMTM{}}
         \label{fig:map_fmtm}
     \end{subfigure}
 \caption{Results, in terms of COCO mAP (\%), obtained for different object-detection models in different scenarios, in which the number of available original documents is scarce. The ``Non-augmented'' bars indicate the results obtained with only original images, whereas the ``Our augmentation'' bars represent the cases in which our data-generation algorithm is used to build 100 synthetic images.}
 \label{fig:results_map}
\end{figure*}

The models are evaluated in terms of COCO \ac{mAP} in different situations of data availability with the aim of studying their behavior according to the number of annotated pages used to train them. Because of the cost of manually annotating music manuscripts, it is particularly relevant to analyze their behavior when limited annotated data is provided. We, therefore, also study the benefits of the algorithm proposed in Section~\ref{sec:method:algorithm} as regards building semi-synthetic images and increasing the number of pages and variability of data. Figure~\ref{fig:results_map} provides a graphic representation of this metric in order to study the effectiveness of each model according to the number of real pages used to train them. It also shows the results obtained after the application of our data augmentation proposal when compared with the use of only the original images. 

First note that the results are, in general, quite modest, mainly because of the rigorousness of the metric used. Moreover, although the amount of pages is crucial as regards optimizing \ac{mAP}, a higher number of pages does not guarantee good results, depending on the difficulty of the corpus. Indeed, \FMT{} attains more overlapping and density of bounding boxes, especially in the case of \FMTC{}, which considerably increases the difficulty of the predictions and which translates into worse detection quality. 

Despite this, it will be observed that, as expected, there is a similar trend for almost all the models, in which the fewer the number of actual pages, the worse the detection gets, since the models do not have sufficient reference data with which to learn patterns in order to generalize the detection. Nevertheless, when our data-augmentation algorithm is applied, these results are drastically improved, obtaining models that are more robust to the lack of ground-truth data.

Focusing on \RetinaNet{} and \FasterRCNN{}, according to the \ac{mAP} results, these are the two models that most benefit from our data-augmentation algorithm for all the corpora considered, since they obtain better results for almost all the scenarios, particularly in those cases in which there is a scarcity of data. The results show an increase in stability for these models, since they achieve more robustness, especially in those cases in which the data are limited. The results for \SAE{} and \SSD{} are also improved with our data-augmentation algorithm, but there are fewer cases in which the augmented data are better than the original ones, depending mainly on the corpus considered. These last models would appear to be more sensitive to the overlapping of the regions, since they are the models that attain the worst results as regards \FMTC{}. In this respect, \FasterRCNN{} with data augmentation obtains the best figures in this challenge corpus, although the results do not reach 25\% of \ac{mAP} in either case.

\begin{table}[!ht]
\centering
\caption{Results obtained for object detection in terms of COCO mAP (\%) for scenarios with a different availability of ground-truth data. The figures in bold type indicate the best results obtained for each scenario according to the number of real pages available.}
\label{tab:results_object_map}
\begin{tabular}{lllllllll}
\toprule
\multicolumn{2}{c}{\multirow{2}{*}{\textbf{Model}}} & \multicolumn{7}{c}{\textbf{Available real pages}} \\ \cmidrule(rl){3-9}
\multicolumn{2}{c}{} & 1 & 2 & 4 & 8 & 16 & 32 & 64 \\ \cmidrule(rl){1-9}
\multicolumn{9}{l}{\RetinaNet{}} \\ \cdashlinelr{1-9}
& \NoAugmented{} & \phantom{0}7.4 & \phantom{0}6.6 & \phantom{0}8.0 & 11.3 & 12.2 & 22.4 & 29.6 \\
& \UniformRandom{} & 23.6 & 24.6 & \textbf{25.6} & 25.6 & 26.8 & 26.7 & 25.2 \\ \cmidrule(rl){1-9}
\multicolumn{9}{l}{\FasterRCNN{}} \\ \cdashlinelr{1-9}
& \NoAugmented{} & \phantom{0}1.6 & \phantom{0}0.8 & \phantom{0}0.8 & 0.1 & 3.8 & 14.8 & 26.3 \\
& \UniformRandom{} & 23.6 & 24.6 & \textbf{25.6} & 25.6 & 26.8 & 26.7 & 25.2 \\ \cmidrule(rl){1-9}
\multicolumn{9}{l}{\SSD{}} \\ \cdashlinelr{1-9}
& \NoAugmented{} & \phantom{0}1.8 & \phantom{0}1.5 & \phantom{0}1.2 & \phantom{0}0.7 & \phantom{0}4.1 & \phantom{0}3.9 & 10.5 \\
& \UniformRandom{} & 10.8 & 12.5 & 12.6 & 12.6 & 15.9 & 14.0 & 17.6 \\ \cmidrule(rl){1-9}
\multicolumn{9}{l}{\SAE{}} \\ \cdashlinelr{1-9}
& \NoAugmented{} & \phantom{0}1.2 & \phantom{0}2.1 & 13.6 & \phantom{0}0.0 & 19.8 & 35.3 & \textbf{37.8} \\
& \UniformRandom{} & \textbf{23.8} & \textbf{26.0} & 21.4 & \textbf{27.3} & \textbf{34.2} & \textbf{34.1} & 31.9 \\
%   \cmidrule(rl){1-9}\morecmidrules\cmidrule(rl){1-9}
%\multicolumn{9}{l}{\textsc{Average}} \\ \cdashlinelr{1-9}
%   & \NoAugmented{} & \phantom{0}3.0 & \phantom{0}2.7 & \phantom{0}5.9 & \phantom{0}3.0 & 10.0 & 19.1 & 26.1 \\
%   & \UniformRandom{} & 22.0 & 23.1 & 22.7 & 24.0 & 26.8 & 26.4 & 26.6 \\ 
\bottomrule
\end{tabular}
\end{table}

Table \ref{tab:results_object_map} shows the average results for all corpora and for each object-detection model for analysis purposes. The results show that no augmentation is not feasible in those cases in which few training pages are available, since poor figures are obtained. This table also shows that \SAE{} achieves the best \ac{mAP} for all the scenarios except that of 4 actual pages, and that in this case, \RetinaNet{} and \FasterRCNN{} obtain a better value of 25.6\% when compared to the 21.4\% obtained by \SAE{}. An interesting point when comparing the cases with and without image augmentation is that, on average, all the models obtain more stable figures in all cases. This is particularly interesting because it justifies the need for an algorithm with which to obtain more robust models, and this improvement is especially noteworthy when few pages are available to train them. 

In either case, this experiment makes it possible to conclude that, on average, the model that should be used for \ac{LA}, at least according to \ac{mAP}, is \ac{SAE} with our augmentation algorithm. The suitability of this metric as regards representing the retrieval of regions is analyzed in the case study shown below, since it is the most important factor for a successful \ac{LA}.

\subsection{Case Study II: Evaluation in terms of regions retrieved }
\label{sec:results:second_experiment}

It should be noted that \ac{mAP} is a metric that measures the robustness of the models in terms of probabilities and overlapping, and is not a discrete object retrieval metric. With regard to \ac{OMR}, for which the importance of \ac{LA} lies in the detection of regions, and not so much in the degree of overlapping with the ground-truth regions, this metric may not correctly represent the number of regions retrieved. In this case study, we discuss the convenience of the popular metric \ac{mAP} when used in object detection and explore whether it is aligned with a greater number of detected regions, which is really the main goal of region-based \ac{LA}. As mentioned above, this can be measured using \MacroPrecision{}, \MacroRecall{} and \MacroFscore{}, and we, therefore, compare the conclusions extracted from the previous case study with those obtained by means of these metrics.

However, in order to consider that a predicted region has been corrected, it is necessary to define two thresholds: one for confidence and another for \ac{IoU}. Confidence, as mentioned in Section~\ref{sec:method:architectures}, is the level of certainty that the object-detection model has when predicting the bounding boxes, whereas \ac{IoU} indicates the degree of overlapping between the predicted regions and the real ones. The results obtained for a region considered to be a correct prediction should, therefore, have sufficient confidence and \ac{IoU}, signifying that those predicted regions that do not surpass these thresholds are discarded. For this reason, and because of their importance, exhaustive experimentation has been performed to obtain the best combination in the validation set. For the confidence threshold, we considered a range of values of between 0.05 and 0.95 with intervals of 0.1, while in the case of \ac{IoU}, we explored common values used in object detection, specifically between 0.5 and 0.95 with a granularity of 0.05. 

We subsequently considered \MacroPrecision{}, \MacroRecall{} and \MacroFscore{} in order to evaluate the results. Table~\ref{tab:results_object_fscore} shows the average results obtained for the best combination of thresholds in each case after optimizing \Fscore{}. Note that confidence is a value that could be used in practice, since it is provided by the object detection model, but that the IoU can be used only in controlled scenarios, since it is computed by using the ground-truth data. The combination of both thresholds, therefore, provides a reference of the best results that could be obtained.

\begin{table}[!ht]
\centering
\caption{Average results in terms of \MacroPrecision{}, \MacroRecall{} and \MacroFscore{} (\%). Figures in bold type represent the best values for each metric and for each scenario considered, i.e., for a different number of available real pages. Underlined values indicate the best results for each metric, considering all the cases.}
\label{tab:results_object_fscore}
\renewcommand{\arraystretch}{0.4}
\centering
\resizebox{0.85\textwidth}{!}{
\begin{tabular}{lllllllll}
\toprule
\multicolumn{2}{c}{\multirow{3}{*}{\textbf{Scenarios}}}  & \multicolumn{7}{c}{\textbf{With augmentation}} \\ 
\cmidrule(rl){3-9} 
\multicolumn{2}{c}{} & \multicolumn{3}{c}{No} & \multicolumn{1}{c}{} & \multicolumn{3}{c}{Yes} \\ 
\cmidrule(rl){3-5} \cmidrule(rl){7-9} 
\multicolumn{2}{c}{} & \multicolumn{1}{c}{\MacroPrecision{}} & \multicolumn{1}{c}{\MacroRecall{}} & \multicolumn{1}{c}{\MacroFscore{}} & \multicolumn{1}{c}{} & \multicolumn{1}{c}{\MacroPrecision{}} & \multicolumn{1}{c}{\MacroRecall{}} & \multicolumn{1}{c}{\MacroFscore{}} \\ 
%\cmidrule(rl){1-10}
\midrule
\multicolumn{2}{l}{1 page}&&&&&&&\\ \cdashlinelr{1-9}
& \RetinaNet{} & \phantom{0}7.2&14.2& \phantom{0}9.6&& 61.9 & \textbf{62.6} & 62.3 \\
& \FasterRCNN{}& \phantom{0}5.2& 31.0& \phantom{0}8.9&& 71.1 & 60.5 & \textbf{65.3} \\
& \SSD{}& \phantom{0}5.3& 20.5& \phantom{0}8.4&& 56.8 & 46.6 & 51.2 \\
& \SAE{}& \underline{\textbf{100}} & \phantom{0}2.9& \phantom{0}5.7&& 62.1& 28.0 & 38.6 \\ \cmidrule(rl){1-9} 
\multicolumn{2}{l}{2 pages}&&&&&&&\\ \cdashlinelr{1-9}
& \RetinaNet{} & \phantom{0}6.4& 12.8& \phantom{0}8.5&& 53.7 & \textbf{70.6} & 61.0 \\
& \FasterRCNN{}& 45.2& 25.1& 32.3&& \textbf{80.9} & 51.6 & \textbf{63.0} \\
& \SSD{}&71.4& 17.6& 28.2&& 53.2& 46.0 & 49.3 \\
& \SAE{}& 18.1& \phantom{0}6.6& \phantom{0}9.7&& 66.4 & 38.2 & 48.5 \\ \cmidrule(rl){1-9} 
\multicolumn{2}{l}{4 pages}&&&&&&&\\ \cdashlinelr{1-9}
& \RetinaNet{}& \phantom{0}6.2& 32.8& 10.4&& \textbf{76.5} & 55.2 & \textbf{64.1} \\
& \FasterRCNN{}& 30.2& 36.0& 32.8&& 54.8 & \textbf{69.6} & 61.3 \\
& \SSD{}& 61.3& 11.8& 19.8&& 59.9& 38.7 & 47.7 \\
& \SAE{}& 36.4& 23.3& 28.4&&38.2 & 32.8 & 35.3 \\ \cmidrule(rl){1-9} 
\multicolumn{2}{l}{8 pages}&&&&&&&\\ \cdashlinelr{1-9}
& \RetinaNet{} & 32.2& 40.0& 35.7&& 61.8 & \textbf{71.6} & \textbf{66.4} \\
& \FasterRCNN{}& \phantom{0}6.7& \phantom{0}7.9& \phantom{0}7.2&& \textbf{66.6} & 66.0 & 66.3 \\
& \SSD{}& 31.2& \phantom{0}6.1& 10.2&& 64.9 & 23.4 & 34.4 \\
& \SAE{}& \phantom{0}0.0& \phantom{0}0.0& \phantom{0}0.0&& 40.9 & 36.1 & 38.3 \\ \cmidrule(rl){1-9} 
\multicolumn{2}{l}{16 pages}&&&&&&&\\ \cdashlinelr{1-9}
& \RetinaNet{} &\textbf{67.9}& 41.9& 51.8&& 60.2& \underline{\textbf{77.6}} & \textbf{67.8} \\
& \FasterRCNN{}& 51.1& 31.1& 38.7&& 65.0 & 69.6 & 67.2 \\
& \SSD{}& 52.3& 28.8& 37.2&& 42.9& 42.2 & 42.6 \\
& \SAE{}& 42.5& 31.1& 35.9&&50.7 & 55.0 & 52.8 \\ \cmidrule(rl){1-9} 
\multicolumn{2}{l}{32 pages}&&&&&&&\\ \cdashlinelr{1-9}
& \RetinaNet{}& 66.1& 58.6& 62.1&& 56.0& \textbf{73.4} & 63.6 \\
& \FasterRCNN{}& 43.4& 49.4& 46.2&& \textbf{68.9} & 63.2 & \textbf{65.9} \\
& \SSD{}& 67.9  & \phantom{0}9.4& 16.5&& 45.5& 45.5 & 45.5 \\
& \SAE{}& 53.5& 54.8 & 54.1 && 54.6 & 51.7& 53.1\\ \cmidrule(rl){1-9} 
\multicolumn{2}{l}{64 pages}&&&&&&&\\ \cdashlinelr{1-9}
& \RetinaNet{} & \textbf{93.4}& 54.8& 69.1&& 75.7& \textbf{68.0} & \underline{\textbf{71.6}} \\
& \FasterRCNN{}& 61.1& 64.6 & 62.8&& 75.1 & 57.7& 65.2 \\
& \SSD{}& 75.0& 29.6& 42.5&& 65.9& 55.1 & 60.0 \\
& \SAE{} &53.6&57.9&55.7 && 51.8& 51.0& 51.4\\ 
\bottomrule
\end{tabular}
}
\end{table}

\begin{figure*}[!ht]
     \centering
     \begin{subfigure}[b]{1.0\textwidth}
         \centering
         \includegraphics[width=\textwidth]{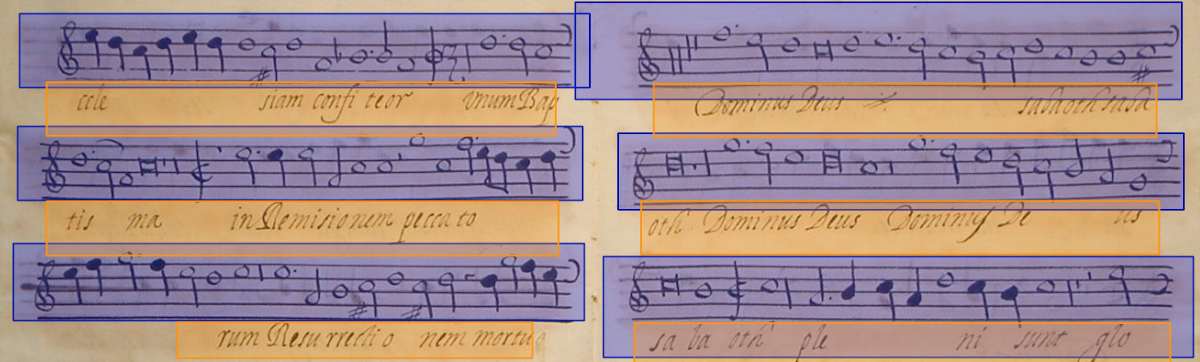}
         \caption{\Capitan{}}
         \label{fig:qualitative:capitan}
     \end{subfigure}

     \begin{subfigure}[b]{1.0\textwidth}
         \centering
         \includegraphics[width=\textwidth]{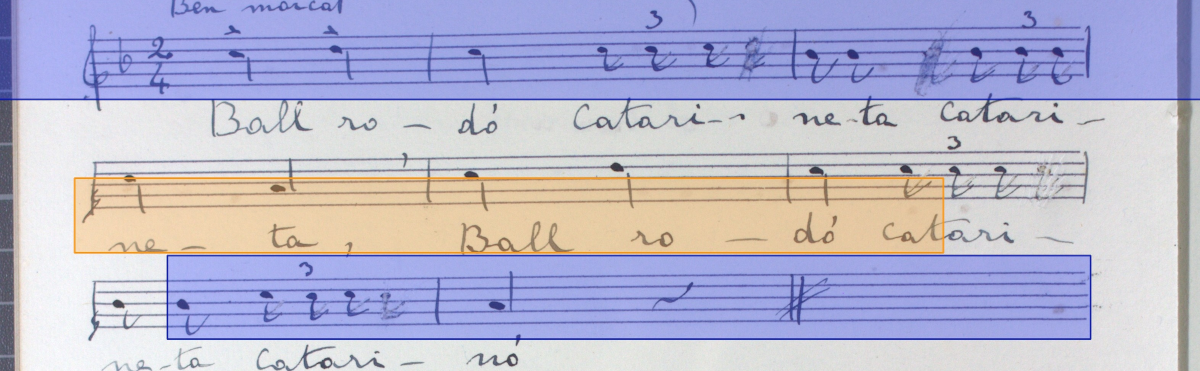}
         \caption{\FMTC{}}
         \label{fig:qualitative:fmt}
     \end{subfigure}
     \caption{Selected extractions of object detection in which correct and incorrect estimations are shown by using \RetinaNet{}. Blue bounding boxes represent predictions of staff regions, whereas the orange boxes represent the lyric areas obtained.}
     \label{fig:qualitative}
\end{figure*}

It will first be noted that \ac{mAP} and \MacroFscore{} do not match in the model with the best figures. As shown in Table~\ref{tab:results_object_map}, in the case of \ac{mAP}, the highest value was obtained by \SAE{}, with an average of 37.8\% for 64 original pages and non-augmentation. This model also achieved the best performance for most of the augmentation scenarios, making \SAE{} the best model in general according to \ac{mAP}. However, as reported in Table~\ref{tab:results_object_fscore}, the \MacroFscore{} metric indicates that the best results are provided by \RetinaNet{} in combination with our data-augmentation algorithm when 64 real pages are available, with a \MacroFscore{} of 71.6\%. Moreover, most of the augmented scenarios indicate that \RetinaNet{} is the best option, followed by \FasterRCNN{} -  two different models to those stated by \ac{mAP}. What is more, the \MacroFscore{} figures are considerably higher than the \ac{mAP} figures, and, as will be seen later in Figure~\ref{fig:qualitative}, visually, higher values are more correlated with the regions retrieved. %This analysis, therefore, shows that \Fscore{} is a more appropriate metric with which to measure the performance of \ac{LA} in \ac{OMR}.

Focusing on the results, it will be observed that the augmentation algorithm proposed is crucial in terms of \MacroFscore{}. In none of the non-augmented cases does this metric supersede the augmented scenarios, and this, therefore, justifies the theory that this algorithm is able to increase the robustness of the models as regards extracting the bounding boxes. There are some examples as regards the non-augmented experiment in which \MacroPrecision{} improves the results obtained after augmentation, particularly in the case of \SAE{} with 1 page, which obtains 100\%, while \RetinaNet{} yields 67.9\% and 93.4\% for 16 and 64 pages, respectively. However, when focusing on the values of \MacroRecall{} for these cases, it will be noted that \SAE{} attains only 2.9\%, and \RetinaNet{} obtains 41.9\% and 54.8\%, respectively. This signifies that, in these cases, the models prioritize the detection of real regions over the miss-detection of regions in which there is no information. In other words, there are regions that have to be manually discarded.  This situation could be interesting depending on the task being carried out, but a balanced model could, in general terms, be beneficial as regards obtaining good results with less human intervention.

As shown in Table~\ref{tab:results_object_fscore}, the augmented scenarios attain more stable and balanced results in terms of \MacroPrecision{} and \MacroRecall{}, impacting directly on better \MacroFscore{} figures. Indeed, for this metric, all the experiments evaluated with augmentation were, on average better than the non-augmented cases. According to these metrics, the best models for \ac{LA} are \RetinaNet{} and \FasterRCNN{}, with \RetinaNet{} being a potential solution since it obtains the highest \MacroFscore{} value---71.6\%---with 64 pages, and the best \MacroRecall{} with a value of 77.6\%. 

However, in the experiment shown previously in Section~\ref{sec:results:first_experiment}, \SAE{} is the best model, and the conclusions for both experiments are, therefore, different. It is precisely this situation which justifies that the \ac{mAP} metric is not suitable for evaluating \ac{LA}, since the real importance of this process is the extraction of bounding boxes as objects, and not only the evaluation of their overlapping.

In order to complement the \ac{LA} experiments, Figure~\ref{fig:qualitative} shows an example from \Capitan{} in which the bounding boxes are correctly predicted and another one in which there are several miss-detections. 

In the first example, which is shown in Figure~\ref{fig:qualitative:capitan}, it will be observed that the retrieval region is generally of good quality. In this example, the bounding boxes retrieved appear to correctly contain the relevant information, but, it should be noted that the staff retrieval in this example obtains an average \ac{IoU} of 79\% and a confidence of 55\%, whereas the text retrieval obtains an \ac{IoU} of  74\% and a confidence of 39\%. That signifies that, although the prediction of the bounding would appear to be graphically suitable and correct, since the objects are large and regular, a slight error, especially on the vertical side in our context, may considerably worsen the \ac{IoU}. This, therefore, means that it is not necessary to attain a perfect matching of \ac{IoU} in order to cover the data that has to be retrieved, and this also explains why the \ac{mAP} obtains significantly low figures, since the range between 80\% to 95\% would not, on average, contain any bounding boxes. Moreover, the confidence of the model is, on average, very poor when compared to what might be expected after visual inspection, signifying that a particularly low confidence threshold is needed for this metric in order to prevent these regions from being discarded.

In the second example depicted in Figure~\ref{fig:qualitative:fmt}, there are certain issues as regards both staff and text retrieval. In visual terms, the staff at the top appears to have been correctly retrieved, although the area detected is higher than the staff itself. The principal problem with the staves in this case is that one staff is not detected, the last one is partially retrieved, and two music symbols are missed. These issues are crucial for the eventual transcription, and a manual correction would, therefore, be required in order to correctly extract the bounding boxes. With regard to the text regions, only one bounding box is retrieved, but it does not cover the text at all. Two other text lines are not detected, and manual corrections should, therefore, also be performed for a full digitization. In this example, the staff predictions have 57\% of average \ac{IoU} with respect to the ground truth, whereas \RetinaNet{} provides an average confidence of only 28\%. In the case of the text regions, the \ac{IoU} obtained is 25\%, despite the fact that the text is quite well detected, and the confidence reaches 59\%. 

This qualitative analysis, therefore, reinforces the idea that it is not necessary to obtain a perfect matching of the bounding boxes, and that obtaining regions that encompass the content is sufficient. This demonstrates that \ac{mAP} is not an appropriate metric with which to evaluate the objects retrieved, since it places much more importance on the overlapping with the ground truth. It is consequently possible to conclude that the \MacroFscore{} metric is more suitable than \ac{mAP} for the \ac{LA} of music score images, despite the popularity of \ac{mAP} in object detection. To complete and confirm our analysis, in the next case study, we further analyze the influence of \ac{IoU} and confidence in the final transcription.

\subsection{Case Study III: Goal-directed evaluation}
\label{sec:results:third_experiment}
We have, until this point, performed a thorough analysis of the \ac{LA} stage on its own, without any specific context. However, it is important to recall that, in most cases, this stage is not an objective in itself, but merely an intermediate step within a pipeline employed to transcribe the content of music score images. In this section, therefore, we study the relationship between the operation of \ac{LA} and the transcription process itself, focusing particularly on the music notation (regions with staves). To this end, we selected \RetinaNet{} as being representative of an automatic layout analysis stage, given that it was, according to our previous experiment, the best option.

For this goal-directed experiment, we employed a state-of-the-art model for \ac{OMR} that is built as a \ac{CRNN} and is directly trained to retrieve the sequence of musical symbols found in the image of a single staff. Since the \ac{CRNN} is used here as a black box, the reader is referred to a number of works for further details on its operation \citep{shi2016end,Calvo-ZaragozaT19, wick2021experiments}.

The experiment outlined in this section is as follows:
\begin{enumerate}
    \item For each corpus, and using the training and validation partitions, the \ac{CRNN} is trained by means of the ground-truth regions along with their corresponding transcripts, thus ensuring that the recognition model is the best possible.
    \item  With regard to the test partitions, we employ the \ac{LA} model to automatically retrieve the staff regions, along with their confidence.
    \item Each predicted staff is matched with all the ground-truth regions of the test partition for which the \ac{IoU} is greater than $0.55$.
    \item For each match, both the detected and the ground-truth staves are processed with the \ac{CRNN} in order to retrieve their music symbols. 
\end{enumerate} 

We denote as \diffSER{} the difference in SER between the symbols retrieved from the ground-truth staff and the symbols retrieved from the detected staff. This will be used as a measure of the impact of the layout analysis: if $0$, this signifies that there is no actual difference between retrieving the content using the manually-annotated region and retrieving the content using the automatically-detected region (a fairly ideal scenario). As this difference grows, the performance loss caused by the layout analysis is greater. In turn, it might occur that the SER is smaller in the region predicted automatically, signifying that the \diffSER{} would be negative. Whatever the case may be, for each detected region, we eventually obtain a tuple (\diffSER, confidence, \ac{IoU}).

Furthermore, before reporting the results of this experiment, it should be taken into account that some deviations in the detected regions could be alleviated by training the CRNN with data augmentation by, for example, slightly modifying the corners of the training staff regions. The effect of data augmentation on staff-based OMR with CRNN has already been studied in previous works~\citep{juanki2021}, although not comprehensively in the context of its connection with an imperfect layout analysis. Here we shall, therefore, consider the CRNN with and without this type of data augmentation in order to also carry out the study from this perspective.

\begin{figure*}[!ht]
     \centering
     \begin{subfigure}[b]{0.49\textwidth}
         \centering
         \includegraphics[width=\textwidth]{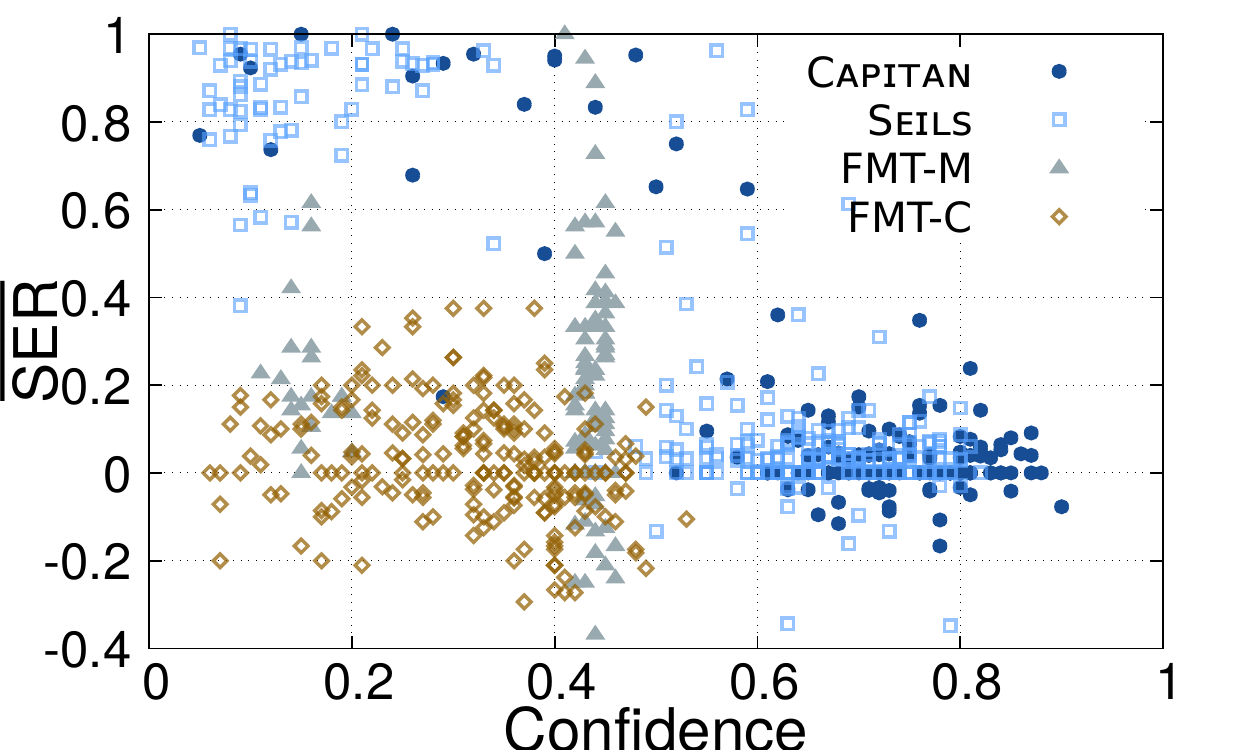}
         \caption{Confidence vs \diffSER{}.}
         \label{fig:CvS_no}
     \end{subfigure}
     \begin{subfigure}[b]{0.49\textwidth}
         \centering
         \includegraphics[width=\textwidth]{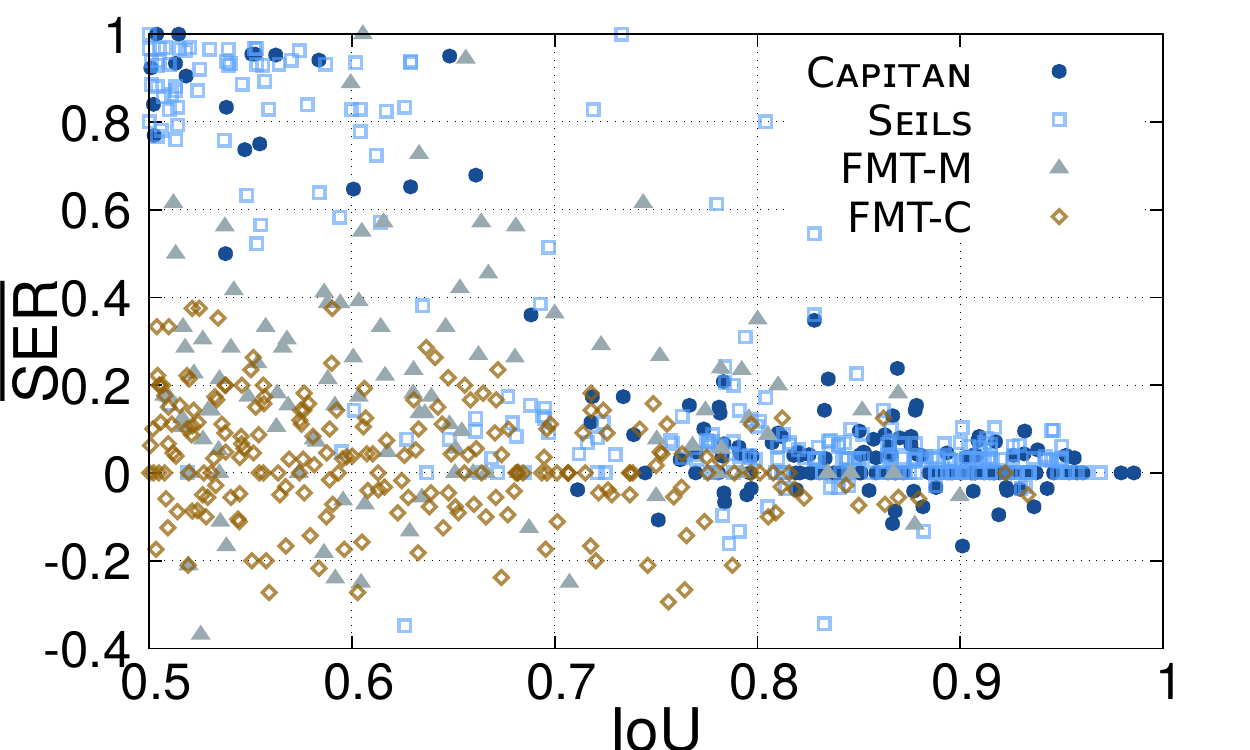}
         \caption{\ac{IoU} vs \diffSER{} without augmentation}
         \label{fig:IvS_no}
     \end{subfigure}
     \begin{subfigure}[b]{0.49\textwidth}
         \centering
         \includegraphics[width=\textwidth]{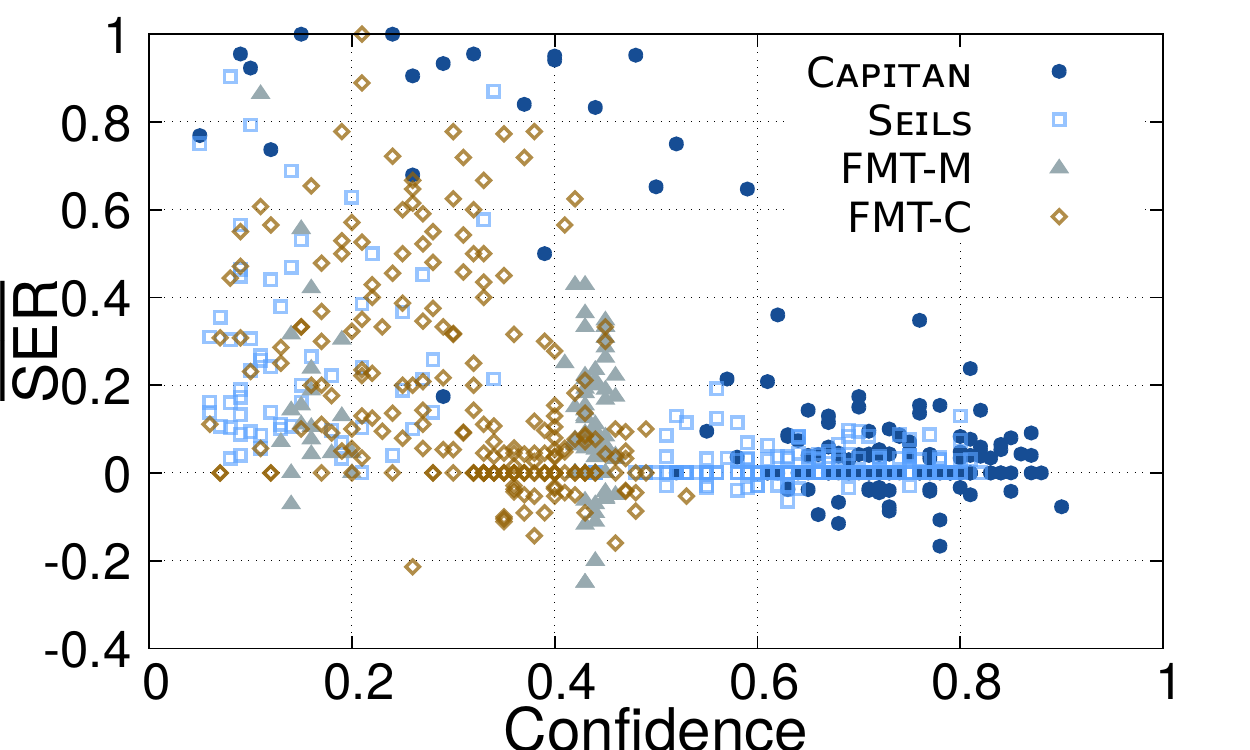}
         \caption{Confidence vs \diffSER{} with augmentation.}
         \label{fig:CvS_yes}
     \end{subfigure}
     \begin{subfigure}[b]{0.49\textwidth}
         \centering
         \includegraphics[width=\textwidth]{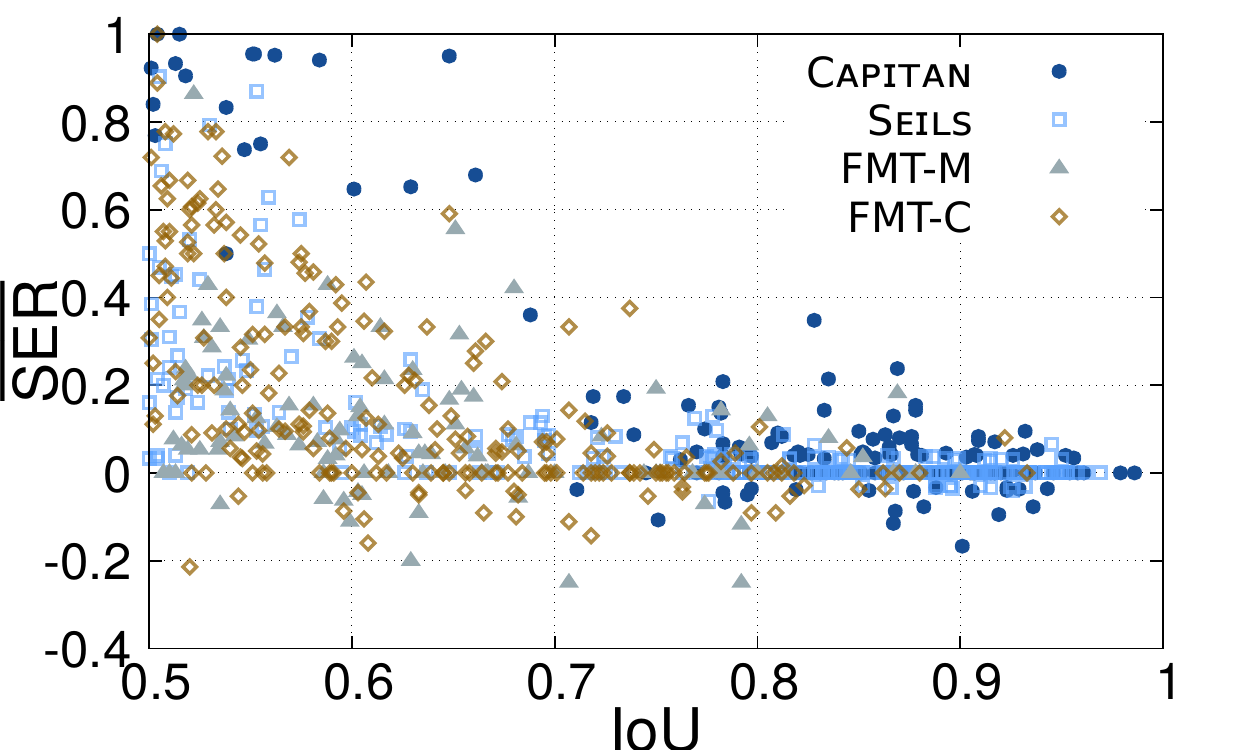}
         \caption{\ac{IoU} vs \diffSER{} with augmentation}
         \label{fig:IvS_yes}
     \end{subfigure}
     \caption{Relationship between IoU and confidence obtained for \ac{LA} and the music transcription, in this case scored using \diffSER{}. }
     \label{fig:results_ser}
\end{figure*}

Figure \ref{fig:results_ser} shows the contrast of the confidence and \ac{IoU} values ($x$-axes) of the detected regions with the \diffSER{} ($y$-axes), highlighting the different corpora. An initial remark is that, as might be expected, the \ac{LA} has less impact in the regions that have a higher confidence and a higher \ac{IoU} (right part of the images), in which the \diffSER{} is closer to 0. As the model has less confidence in the regions or the \ac{IoU} decreases, this value clearly increases. This even produces cases of \diffSER{} = $1$, signifying that the \ac{CRNN} perfectly retrieves the symbols for the ground-truth region but completely fails in the case of the predicted one. 

Particularly in the case of confidence (Figs. \ref{fig:CvS_no} and \ref{fig:CvS_yes}), the aforementioned phenomenon has a double reading: while it is true that the results are quite poor when transcribing the staff from the less reliable regions, these could easily be discarded. The full \ac{OMR} system should consider only the regions for which the confidence is high and for which a positive result is, in most cases, expected. For this \ac{LA} model, and for all corpora in general, it appears that a suitable threshold for such a purpose would be $0.6$. 

Furthermore, the correlation between the \ac{IoU} and the \diffSER{} also produces a clear trend (Figs. \ref{fig:IvS_no} and \ref{fig:IvS_yes}): the higher the \ac{IoU}, the lower the deterioration of the transcription. Unlike confidence, this case cannot be predicted in practice, since the \ac{IoU} can be computed only in controlled experiments in which the true bounding box of a region is known. However, these results could serve to better validate the models in training time. In this case, the threshold beyond which the results drastically change the \diffSER{} depends on the \ac{CRNN} that is used, as discussed below. 

If we compare the results of the base \ac{CRNN} (Figs. \ref{fig:CvS_no} and \ref{fig:IvS_no}) with a \ac{CRNN} trained with data augmentation for the regions (Figs. \ref{fig:CvS_yes} and \ref{fig:IvS_yes}), it is clear that the latter is much more robust to an (imperfect) automatic \ac{LA}. In the case of confidence, there is not much difference; however, in the case of \ac{IoU}, the results are notably better. While without data augmentation, the threshold from which the results are reliable is around $0.9$--- signifying that an almost perfect match is required---the data augmentation manages to enable the \ac{CRNN} to correctly recover the musical symbols with an \ac{IoU} of above approximately $0.7$,  signifying a much more reasonable case to attain in practice.

\section{Conclusions}
\label{sec:conclusions}
This work presents comprehensive experiments carried out in order to assess the region-based \ac{LA} process for music score images. This was done by carrying out three specific case studies in which different aspects and goals were assessed. 

The first case study focused on an analysis of the behavior of several well-known object-detection models in different scenarios according to the availability of ground-truth data, which are often scarce. In order to palliate this situation when few annotated images are provided, we proposed and evaluated a data-augmentation algorithm with which to generate semi-synthetic images from the bounding boxes of the original pages. In this case study, we considered a common metric used in object detection in multiple contexts: the \ac{mAP}. The results obtained when employing this metric suggest that the framework based on \ac{SAE} is the best option as regards extracting the different regions, although \RetinaNet{} and \FasterRCNN{} also obtain competitive \ac{mAP} figures, whereas the model with the lowest performance when this metric is employed is \SSD{}.

The objective of the second case study was to demonstrate that the metric considered previously---\ac{mAP}---is not necessarily the best means of evaluating models for the \ac{LA} of music score images. This metric addresses the assessment as an overlapping problem; however, in \ac{OMR}, the number of predicted regions considered as being correct is even more crucial than the overlap between the predicted and the real bounding boxes, as long as the relevant information is included within these regions. We, therefore, proposed an evaluation by using the macro average versions of precision, recall and f-score---\MacroPrecision{}, \MacroRecall{} and \MacroFscore{}, respectively. After the analysis, the model that obtained the highest \MacroRecall{} and \MacroFscore{} was \RetinaNet{}, with 77.6\% and 71.6\%, respectively, and it generally attained more stable and balanced figures for all the models. This also proves that \ac{mAP} is not an appropriate metric for this task.

With regard to the third case study, we attempted to evaluate the relationship between the overlapping of predicted and annotated staff regions, the confidence provided by the \ac{LA} model, and the error obtained in the final transcription through the use of an end-to-end strategy by means of \ac{CRNN}, measured with the \ac{SER} metric. We additionally explored the influence of data augmentation shown in previous works on these relationships. One of the main conclusions obtained was that high confidence and \ac{IoU} values are strongly aligned with low transcription errors. Indeed, in the case of all the corpora evaluated, we observed an abrupt reduction in the error from a certain value of confidence and \ac{IoU}. This supports the idea that using thresholds to filter the \ac{LA} regions is a correct way in which to discard those regions that may cause errors in the transcription. This ensures a certain quality of the transcriptions, which could be used to train other end-to-end models.

As the aforementioned results show, no model detects all the regions of interest in music score images. A specific object-detection model for \ac{LA} in \ac{OMR} could be a promising avenue for further research, in which specific characteristics of this type of documents could be exploited. For example, the fact that their general structure is regular or that their regions are usually wider than taller. Furthermore, it would be interesting to evaluate the performance of these models in cross-manuscript cases (a model trained for one collection and used in another) and to propose improvement strategies in this regard using un- or semi-supervised domain adaptation techniques.

\section*{Acknowledgments}
This paper is part of the I+D+i PID2020-118447RA-I00 (MultiScore) project funded by MCIN/AEI/10.13039/501100011033 and the GV/2020/030 project funded by the Generalitat Valenciana. The first and third authors acknowledge support from the ``ProgramaI+D+i de la Generalitat Valenciana'' through grants ACIF/2019/042 and ACIF/2021/356, respectively.

\bibliography{mybibfile}

\end{document}